\documentclass[10pt,journal,cspaper,compsoc]{IEEEtran}
\pdfoutput=1
\usepackage{booktabs} 
\usepackage{epsfig}
\usepackage{graphicx}
\usepackage{amsmath}
\usepackage{amssymb}
\usepackage{algorithm}
\usepackage{algorithmic}
\usepackage{subfigure}
\usepackage{url}
\usepackage{multirow}
\usepackage[usenames,dvipsnames,table]{xcolor}

\newcommand{\tabincell}[2]{\begin{tabular}{@{}#1@{}}#2\end{tabular}}

\begin{document}
%
\title{Deep Human Parsing with Active Template Regression}

\author{Xiaodan~Liang, Si~Liu, Xiaohui Shen, Jianchao Yang, Luoqi Liu, Jian Dong, Liang Lin, Shuicheng Yan,~\IEEEmembership{Senior Member,~IEEE}
\IEEEcompsocitemizethanks{\IEEEcompsocthanksitem Xiaodan Liang is with the School of Information Science and Technology, Sun Yat-sen University, and also with Department of Electrical and Computer Engineering, National University of Singapore. This work was done when the author is intern in the National University of Singapore. Xiaohui Shen and Jianchao Yang are with the Adobe Research, San Jose, California. Liang Lin is with School of Advanced Computing, Sun Yat-sen Unviersity, and also with SYSU-CMU Shunde International Joint Research Institute, Shunde, China.  Luoqi Liu, Jian Dong and Shuicheng Yan are with Department of Electrical and Computer Engineering, National University of Singapore.
Si Liu is with Institute of Information Engineering, Chinese Academy of Sciences, and also with Department of Electrical and Computer Engineering, National University of Singapore. Corresponding author is Si Liu. }\protect\\

\thanks{}}

\markboth{IEEE TRANSACTIONS ON PATTERN ANALYSIS AND MACHINE INTELLIGENCE, VOL. XX, NO. X, X 20XX}
{Shell \MakeLowercase{\textit{et al.}}: Bare Demo of IEEEtran.cls for Computer Society Journals}

\IEEEcompsoctitleabstractindextext{%
\begin{abstract}
\end{abstract}

In this work, the human parsing task, namely decomposing a human image into semantic fashion/body regions, is formulated as an Active Template Regression (ATR) problem, where the normalized mask of each fashion/body item is expressed as the linear combination of the learned mask templates, and then morphed to a more precise mask with the active shape parameters, including position, scale and visibility of each semantic region. The mask template coefficients and the active shape parameters together can generate the human parsing results, and are thus called the structure outputs for human parsing. The deep Convolutional Neural Network (CNN) is utilized to build the end-to-end relation between the input human image and the structure outputs for human parsing.  More specifically, the structure outputs are predicted by two separate networks. The first CNN network is with max-pooling, and designed to predict the template coefficients for each label mask, while the second CNN network is without max-pooling to preserve sensitivity to label mask position and accurately predict the active shape parameters.  For a new image, the structure outputs of the two networks are fused to generate the probability of each label for each pixel, and super-pixel smoothing is finally used to refine the human parsing result. Comprehensive evaluations on a large dataset well demonstrate the significant superiority of the ATR framework over other state-of-the-arts for human parsing. In particular, the F1-score reaches $64.38\%$ by our ATR framework, significantly higher than $44.76\%$ based on the state-of-the-art algorithm~\cite{Yamaguchiparsing13}.


}

\maketitle

\IEEEdisplaynotcompsoctitleabstractindextext

\IEEEpeerreviewmaketitle

\section{Introduction}

With the fast growth of on-line fashion sales, fashion related applications, such as clothing recognition and retrieval~\cite{Yamaguchiparsing13,liu2012street}, automatic product suggestions~\cite{liu2012hi}, have shown huge potential in e-commerce. Among them, human parsing, namely decomposing a human image into semantic fashion/body regions, serves as the basis of many high-level applications, and has drawn much research attention in recent years \cite{Dongparsing13}~\cite{Yamaguchiparsing13}. 

\begin{figure}[t]
\begin{center}
\includegraphics[scale=0.45]{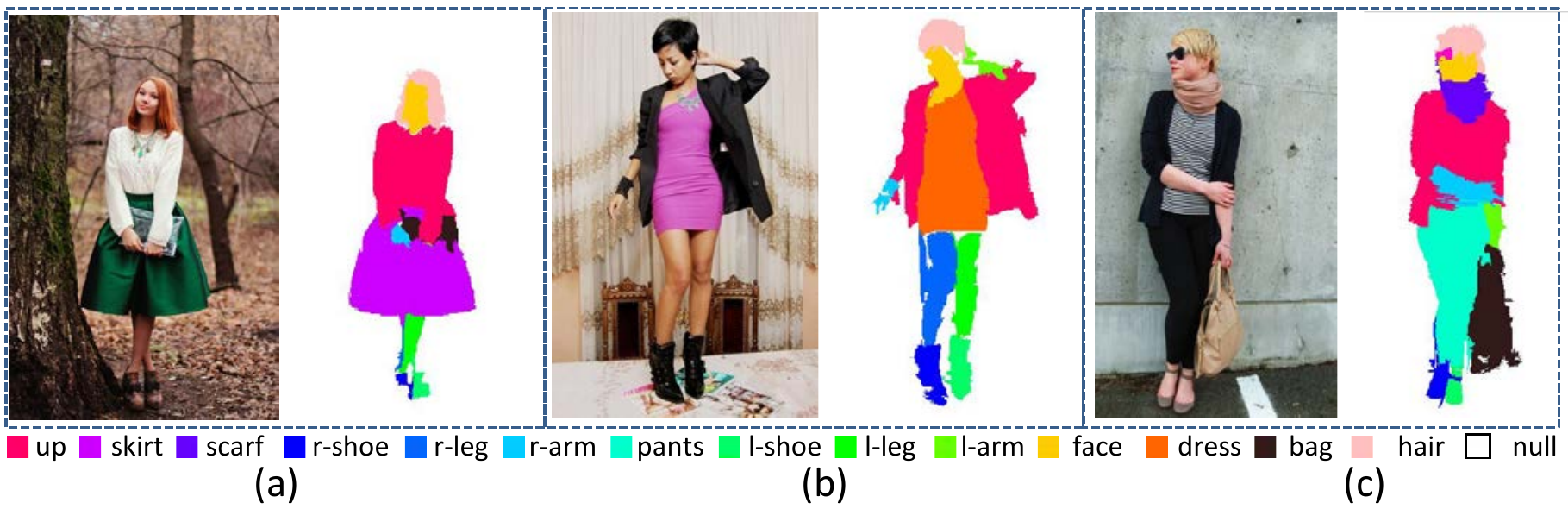}
\vspace{-3mm}
\caption{{Exemplar parsing results by our Active Template Regression (ATR) model. 
For better viewing of all figures in this paper, please see original zoomed-in color pdf file.}}
\label{fig:motivation}
\end{center}
\vspace{-10mm}
\end{figure}

However, there are still some problems with existing algorithms. 
Firstly, some previous works often take the reliable human pose estimation~\cite{dantone2013human} as the prerequisite~\cite{yamaguchi2012parsing}~\cite{Yamaguchiparsing13}~\cite{liu2014fashion2}. However, the possibly bad result from pose estimation shall degrade the performance of human parsing. Secondly, some parsing methods, such as parselets~\cite{Dongparsing13} and co-parsing~\cite{wei14parsing}, which take advantage of the bottom-up  hypotheses  generation methods~\cite{carreira2012cpmc} 
, are implemented based on a critical assumption that the objects or semantic regions have a large probability
to be tightly covered by at least one of the generated
hypotheses. This assumption does not always hold. When the semantic regions appear with larger appearance diversity, it is very difficult to obtain a single hypothesis to cover the whole region, as the object hypotheses by the over-segmentation tend to capture the appearance consistency other than the semantic meanings. 
Thirdly, all existing methods do not sufficiently capture the complex contextual information among the key elements of human parsing, including semantic labels, label masks and their spatial layouts. We argue that human parsing can greatly benefit from the structural information among these elements. As shown in Figure~\ref{fig:motivation}, the presence of the skirt (i.e. its visibility) will hinder the probability of the dress/pants, and meanwhile encourage the visibilities and constrain the locations of left/right legs in (a). For example, the mask of a specific label can also provide the informative guidance for predicting the masks and locations of other labels, especially for the neighboring regions. The mask of the upper-clothes is a single segment due to the presence of the skirt in (c), while the upper-clothes mask is composed of two separate regions due to the dress in (b). Without capturing such structure information, the methods based on low level pixel or region hypotheses are not fully capable of accurately predicting the masks of different labels.

\begin{figure}
\begin{center}
\includegraphics[scale=0.5]{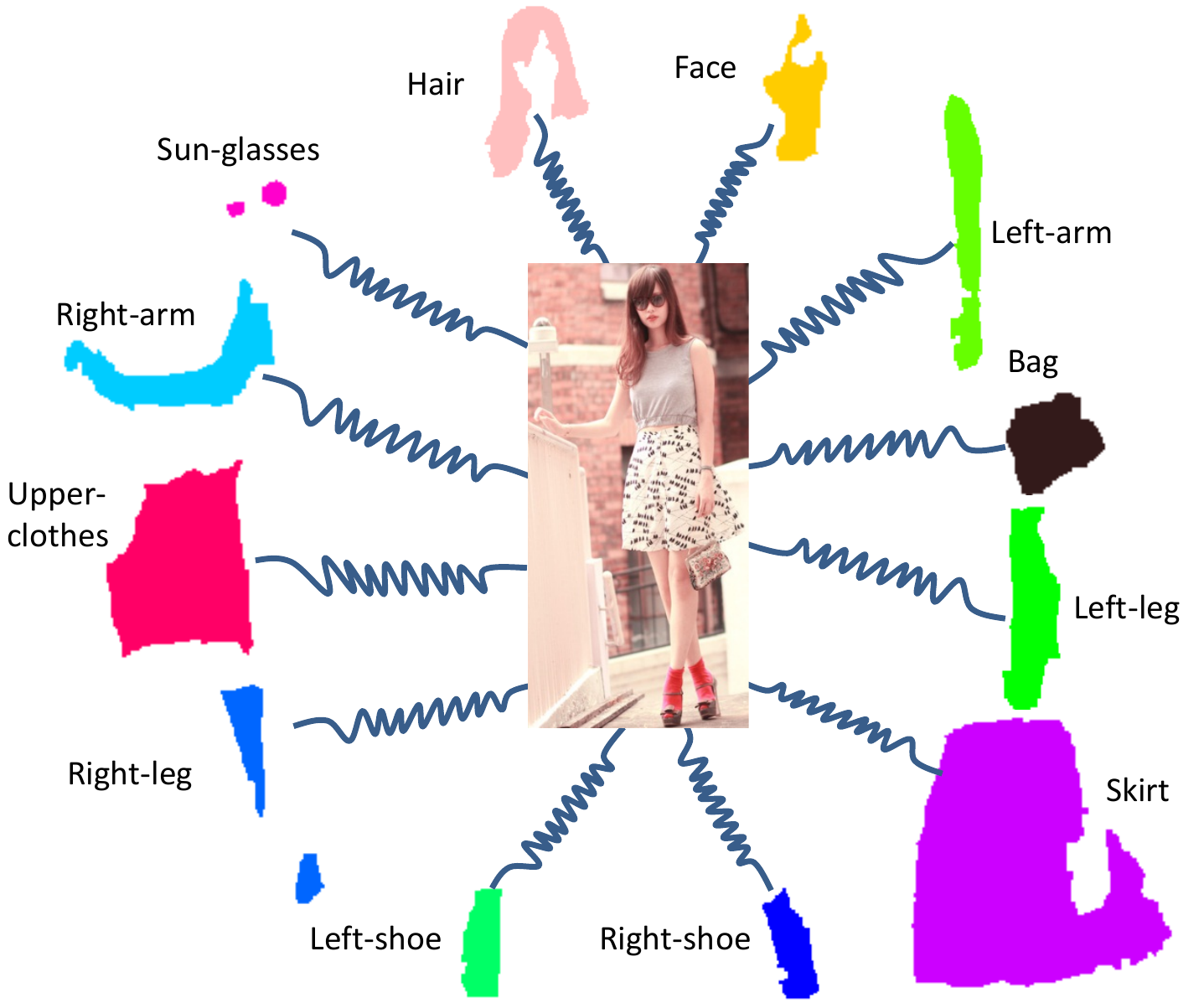}
\vspace{-3mm}
\caption{{Predicted label masks by our model. We directly predict each label mask and then morph them into the absolute image coordinates. 
Different colors indicate different labels. }}
\vspace{-8mm}
\label{fig:starmodel}
\end{center}
\end{figure}

Different from these previous works, we propose a novel end-to-end framework for human parsing and formulate it as an Active Template Regression (ATR) problem. Instead of assigning a label to each pixel or hypothesis, we directly predict and locate the mask of each label. The parsing result for the test image is represented by the set of semantic regions (as in Figure~\ref{fig:starmodel}), which are morphed by the normalized masks with the corresponding active shape parameters, including the position, scale and visibility. In terms of the label mask generation, we first collect all the binary masks of the training images and then learn a batch of mask bases to construct the template dictionary for each label. Intuitively, the template dictionaries can be used to span the subspaces of label masks, which encode the shape priors of each label mask. Any mask with the specific shapes can be generated by adjusting the corresponding template coefficients, inspired by the classic Active Appearance Model (AAM)~\cite{cootes2001active} and Active Shape Model (ASM)~\cite{ASM95}. In this way, our representation is able to capture the natural variability within a set of mask templates for each label. The normalized mask of each label is thus expressed as the linear combination of the mask template dictionary and parameterized by the template coefficients. In terms of active shape parameters, we predict the positions, scales of each semantic region and the visibility flag which indicates whether the specific label appears in the image or not. In this paper, we denote the template coefficients and active shape parameters for each label as two types of structure outputs. Our active template regression framework targets on effectively regressing these structure outputs.

Inspired by its outstanding performance on traditional classification and detection tasks~\cite{deeppose13} \cite{alexnet}  \cite{visualization13}, we utilize the deep Convolutional Neural Network (CNN) to build the end-to-end relation between the input human image and the structure outputs for human parsing, including the mask template coefficients and the active shape parameters. To predict the template coefficients, we aim to find the best linear combination of the learned mask templates. Larger coefficients indicate higher similarities between the label masks and the corresponding templates. The active shape parameters can be predicted similarly as the CNN-based detection task~\cite{deeppose13}. We thus use two separate networks, namely active template network and active shape network, to predict the structure outputs. First, the template coefficients of all labels are together regressed by using the designed active template network which is capable of capturing the contextual correlations among all label masks. Second, the active shape network is designed to predict the position, scale and visibility of each label. To make our active shape network sensitive to position variance, we eliminate the max-pooling layer in the traditional CNN infrastructure~\cite{alexnet}, which is often designed to be invariant to scale and translation changes. For a new photo, the structure outputs of the two networks are fused to generate the probability of each label for each pixel. The super-pixel smoothing is finally used to refine the parsing result. 

To effectively train our networks, we conduct the experiments on a large dataset combining three public parsing dataset and our collected human parsing dataset. Comprehensive evaluations and comparisons well demonstrate the significant superiority of the ATR framework over other state-of-the-arts for human parsing. Furthermore, we also visualize our learned label masks, which demonstrate that our model can generate label masks with strong semantic meanings. Our contributions can be summarized as

\begin{itemize}
    \item Our ATR framework provide an end-to-end approach for human parsing, which directly predicts the label masks and morphs them into the parsing result with active shape parameters. There is no need to explicitly design feature representations, the model topology or contextual interaction among labels.
    \item Our active template network can efficiently predict the most appropriate template coefficients for each label mask, represented by the linear combinations of the template dictionary.
    \item Our active shape network is designed to eliminate max-pooling for accurate position prediction and shows superiority in accurately regressing the active shape parameters over the generic network for classification~\cite{alexnet}.
\end{itemize}

\vspace{-5mm}
\section{Related Work} \label{sec:related_work}
Many research efforts have been devoted into human
parsing. Despite the important
role of human parsing in many fashion-related and human-centric applications~\cite{chen2012describing}~\cite{liu2012street}, it has
not been fully solved. Previous methods are generally based on two
types of pipelines: the hand-designed pipeline and the deep learning pipeline.

\vspace{-5mm}
\subsection{Hand-designed Pipeline}
The traditional pipeline often requires many hand-designed processing steps to perform human parsing, each of which needs to be carefully designed and tuned~\cite{shotton2008semantic} \cite{carreira2012semantic} \cite{yamaguchi2012parsing}\cite{fulkerson2009class}\cite{wei14parsing}~\cite{siparsingTMM14}. These steps use the low-level over-segmentation and pose estimation as the building blocks of human parsing. The classic composite And-Or graph template~\cite{zhu06,lin2014discriminatively} is utilized to model and parse clothing configurations.
 Yamaguchi et al.~\cite{yamaguchi2012parsing} performed human pose estimation and attribute labeling sequentially and then improved clothes parsing with a retrieval-based approach~\cite{Yamaguchiparsing13}. 
 Dong et al.~\cite{Dongparsing13} proposed to use a group of parselets under the structure learning framework. 
However, such  approaches based on hand-crafted relations often fail to fully capture the complex correlations between human appearance and structure. 
Although great progress has been achieved in human parsing, the involved representative model usually requires a lot of prior knowledge about the specific tasks, and these previous methods heavily rely on over-segmentation and pose estimation.

\begin{figure*}
\begin{center}
\includegraphics[scale=0.7]{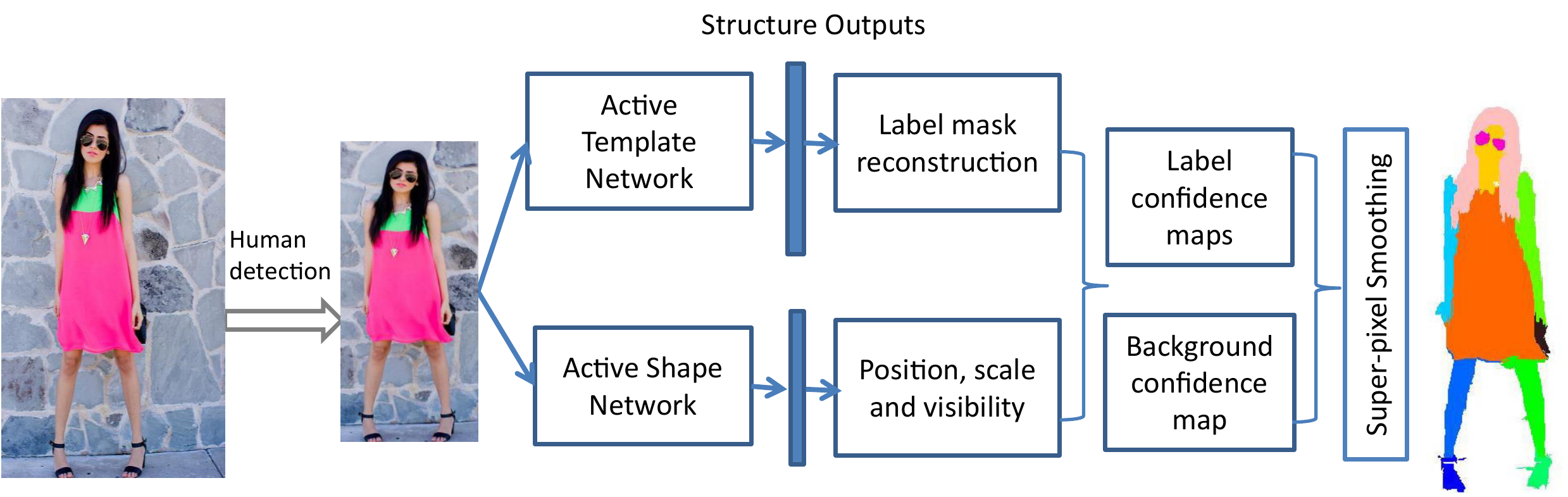}
\vspace{-4mm}
\caption{{Framework of our active template regression model. Given a test image, we first locate the bounding box for human body and then feed it into two separate networks. The template coefficients from the active template network are used to reconstruct the normalized mask. The masks of all labels are then fused together to generate the label confidence maps and the background confidence map by morphing with the active shape parameters (i.e. position, scale and visibility). The super-pixel smoothing is finally used to refine the parsing result. }}
\vspace{-8mm}
\label{fig:framework}
\end{center}
\end{figure*}

\vspace{-4mm}
\subsection{Deep Learning Pipeline}
Recently, rather than using hand-crafted features and model representations, capturing contextual relations and extracting features with deep learning structures, especially deep Convolutional Neural Network (CNN), have shown great potential in various vision tasks, such as image classification~\cite{alexnet}~\cite{visualization13}, object detection~\cite{regionCNN14}, pose estimation~\cite{deeppose13}. To our best knowledge, Convolutional Neural Network has not been applied to human parsing. However, there exist some works on scene parsing and object segmentation with CNN architectures. Farabet et al.~\cite{YannLepami13} trained a multi-scale convolutional network from raw pixels to extract dense features for assigning the label to each pixel. However, multiple complex post-processing methods were required for accurate prediction. The recurrent convolutional neural network~\cite{RecurrentICML14} was proposed to speed up scene parsing and achieved the state-of-the-art performance. Girshick et al.~\cite{regionCNN14} also proposed to classify the candidate regions by CNN for semantic segmentation. All of these approaches use the CNNs as local or semi-local classifiers either over super-pixels or region hypotheses. However, our approach builds an end-to-end relation between the input image and the structure outputs, which is a more efficient application of CNN. 

The above-mentioned hand-crafted and deep models share a similar pipeline: each image is decomposed into small units (pixels, super-pixels or region hypotheses) and local features (hand-crafted features or rich features learned by deep networks) are extracted; then the additional classifiers (shallow models like SVM, or deep models) are trained. In contrast, our approach builds an end-to-end relation between the input image and the structure outputs, which is simple and more efficient. Taking an image as the input, our deep model directly predicts the label masks and the corresponding shape parameters of each semantic region. All the components (e.g., hypothesis generation, feature-extraction and then classification) used in the traditional pipelines are integrated into one unified framework, which distinguishes us from all previous parsing approaches. The closest approaches to ours are~\cite{deeppose13}~\cite{szegedy2013deep} which use CNN-based regression for predicting landmark locations and bounding boxes of the objects, respectively. Their approaches are intuitively similar with our active shape network except that our model eliminates the max-pooling layer for position effectiveness. Moreover, the other important component in our model (i.e. the active template network) is designed to predict the mask template coefficients to actively generate the arbitrary masks of the semantic labels.

\vspace{-3mm}
\section{Active Template Regression}
We formulate the task of human parsing as an active template regression problem. Our framework targets on predicting two kinds of structure outputs: active template coefficients and shape parameters. First, for $K$ different semantic labels (e.g. hair, hat, dress, etc.), we encode the normalized mask of each label as the linear combination of the mask template dictionary $D_k, k = 1,\cdots, K$. Each label mask is parameterized by the corresponding template coefficients, $\boldsymbol{\alpha}_k$, which are treated as the first type of structure outputs. Second, the position of each label mask is parameterized by its top-left coordinates $(b^x_k, b^y_k)\in \mathbb{R}^2$ as well as the width $b^w_k$ and height $b^h_k$. The visibility flag $v_k$ for each label indicates whether the label (e.g. hat, belt) appears in this image. The second type of structure outputs, the active shape parameters, can thus be represented as $\boldsymbol{s}_k = (b^x_k, b^y_k, b^w_k, b^h_k, v_k)$. Finally, the parsing result of the input image $x$ is generated by morphing the masks of all $K$ different labels with the corresponding active shape parameters. In this paper, we train these two types of structure outputs with two separate neural networks: active template network and active shape network, which predict the template coefficients $\{\boldsymbol{\alpha}_k\}_1^K$ and the active shape parameters $\{\boldsymbol{s}_k\}_1^K$, respectively. The reason for training two separate networks is that the learning of template coefficients and shape parameters can be treated as two different tasks: the first one is essentially selecting the most appropriate templates for reconstructing label masks with the template dictionaries, similar to the classification problem, and the second one aims at regressing the precise locations, similar to the detection problem. 

As shown in Figure~\ref{fig:framework}, given an input image, we first detect the human body by using the state-of-the-art detector, i.e., the region convolution neural network method~\cite{regionCNN14}. Considering that the detected bounding box of the human body may not contain all of the body parts, we thus enlarge the detected bounding box with the factor $1.2$. The pixels outside the enlarged bounding box are regarded as the background. The normalized mask of each label is reconstructed by using the predicted template coefficients $\{\boldsymbol{\alpha}_k\}_1^K$ and the template dictionaries $\{D_k\}_1^K$. We then morph these masks into the absolute image coordinates indicated by the shape parameters $\{\boldsymbol{s}_k\}_1^K$. The confidence maps of each label and the background can be obtained according to the morphed masks. Finally, we use the super-pixel smoothing to generate and refine the final parsing result~$y$.

\vspace{-4mm}
\subsection{Active Template Network}
 \label{sec:System_Overview}
 
 \begin{figure*}[t]
 \begin{center}
 \includegraphics[scale=0.6]{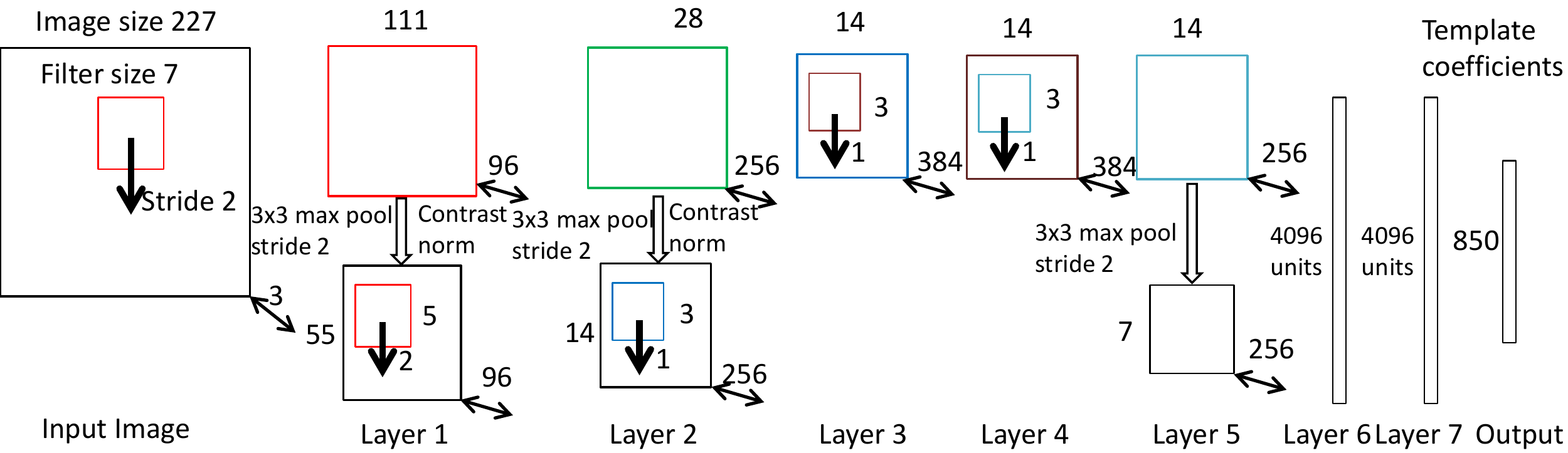}
 \vspace{-3mm}
 \caption{{Our active template network. A $227\times 227\times3$ image is taken as the input. We convolve it with $96$ different 1st layer filters (red), each of which with the size $7\times 7$, using a stride of 2. The obtained feature maps are then: (1) passed through a rectified linear function (not shown), (2) max pooled (within
 $3\times 3$ filter, using stride 2) and (3) contrast normalized. Similar operations are repeated in the 2nd, 3rd, 4th, 5th layers. The last two layers are fully-connected, taking features from the top convolutional layer. The output layer with $850 = 17\times 50$ units is a regression function with $\ell_2$-norm for $K = 17$ labels and each with $M = 50$ coefficients.} }
 \label{fig:activetemplatenet}
 \end{center}
  \vspace{-9mm}
 \end{figure*}
 
The masks of different individual semantic regions for the same label often show various shapes but also common patterns which can distinguish one label from the others. We can thus represent each label mask by the linear combination of the corresponding template dictionary for each label and the label masks can be parameterized by the corresponding template coefficients to best fit the image. Intuitively, the template dictionaries span the subspace  of the label masks and incorporate the shape priors for all labels. By selecting the appropriate template coefficients, we can obtain diverse semantic regions for each label. And the output size of the network can also be significantly reduced by using the template coefficients, rather than using all pixels of the whole mask. 

The active template network is designed to predict the template coefficients. We first generate the mask template dictionaries $\{D_k\}_1^K$ for each label using dictionary learning. More precisely, given the set of training samples $\{x_i,y_i\}_1^n$, we first collect a set of ground-truth binary masks for all $K$ labels. The mask set is denoted by $B_k = \{\boldsymbol{b}_{1,k},\boldsymbol{b}_{2,k},\cdots,\boldsymbol{b}_{n,k}\}$ for the $k$-th label, where $\boldsymbol{b}_{i,k}$ represents the binary mask of the $k$-th label from the sample $(x_i,y_i)$. Specifically, for each label mask, values of the pixels assigned with the specific label are set as $1$ and otherwise  $0$. The binary mask is obtained by the minimum bounding rectangle of the label mask. To learn the template dictionary for each label, we re-scale all these cropped binary masks into a fixed width $r^w$ and height $r^h$. We denote the dictionary for each label as $D_k\in\mathbb{R}^{Z\times M}$ where $Z = r^w\times r^h$, and $M$ as the number of learned templates. The template coefficients of each training sample are denoted as $\boldsymbol{\alpha}_i = \{\boldsymbol{\alpha}_{i,k}\}_1^K$. To jointly predict the template dictionary $D_k$ for each label and the template coefficients $\boldsymbol{\alpha}_{i,k}\in\mathbb{R}^{M}$, we optimize the following cost function for $k$-th label,
 \vspace{-3mm}
 \begin{equation}
 \label{eq:nmf}
 \begin{gathered}
 \min_{\boldsymbol{D_k,\alpha}_{i,k}}\frac{1}{n}\sum_{i=1}^{n}\frac{1}{2}||\boldsymbol{b}_{i,k} - D_k\boldsymbol{\alpha}_{i,k}||_2^2 +\lambda||\boldsymbol{\alpha}_{i,k}||_2^2
  \end{gathered},
 \end{equation} 
  \vspace{-3mm}
  
 \noindent where $\lambda$ is the regularized parameter. It is well-known that $\ell_1$ penalty yields a sparse solution for $\boldsymbol{\alpha}_{i,k}$. However, our active template network with sparse solution may be difficult to converge because of the dominance of the zero values. We thus use the $\ell_2$-norm to regularize the template coefficients. Our experiments demonstrate the superiority with the $\ell_2$-norm than the $\ell_1$-norm. Moreover, we constrain $D_k$ and $\boldsymbol{\alpha}_{i,k}$ to be non-negative, which can help our network generate more reasonable mask templates with semantic meanings than the traditional Principal Component Analysis (PCA)~\cite{pca} methods with both negative and non-negative values~\cite{lee99}. Specifically, the NMF learns part-based decompositions for covering diverse visual patterns of each label and the additive combinations of active templates are beneficial for our reconstruction and network optimization. This Non-negative Matrix Factorization (NMF) problem can thus be effectively solved by the on-line dictionary learning based on stochastic approximations~\cite{nmficml09}.

 We normalize the coefficient values $\boldsymbol{\alpha}_{i,k}$ into the Gaussian distribution with the mean $\boldsymbol{\mu}_k$ and standard deviation $\boldsymbol{\sigma}_k$ for each label. Let $\boldsymbol{\mu}_k = \frac{1}{n}\sum_{i=1}^{n}\boldsymbol{\alpha}_{i,k}$ and $\boldsymbol{\sigma}_k = \sqrt{\frac{1}{n}\sum_{i=1}^{n}||\boldsymbol{\alpha}_{i,k} - \boldsymbol{\mu}_k||^2}$. The normalized temporal coefficients $\hat{\boldsymbol{\alpha}}_{i,k}$ can be defined
 
 \vspace{-3mm}
\begin{equation}
  	\hat{\boldsymbol{\alpha}}_{i,k} = \frac{{\boldsymbol{\alpha}}_{i,k} - \boldsymbol{\mu}_k}{\boldsymbol{\sigma}_k}.
\label{eq:normalize}
\end{equation}
\vspace{-3mm}

We train our active template network to predict the normalized coefficients $\hat{\boldsymbol{\alpha}}_{i}$ based on the Convolutional Neural Network (CNN). The convolutional network consists of several layers and each layer is a linear transformation followed by a non-linear one. The first layer takes an $227\times227\times3$ input image as the input. The last layer outputs the target values of the regression, in our case $M\times K$ dimensions for all labels. Our network is based on the architecture used by Zeiler et al.~\cite{visualization13} for image classification since it has shown better performance on the ImageNet benchmark than the one used by Krizhevsky~\cite{alexnet}. Each layer consists of: (1)~convolution of the
previous layer output (or, in the case of the 1st layer, the image) with a set of filters; (2)~passing the responses through a rectified linear function; (3)~(optionally) max pooling over local neighborhood; (4)~(optionally) the local contrast function that normalizes the responses across feature maps. The top few layers of the network are fully-connected and the final layer is
an $\ell_2$-norm regressor. We refer the reader to Zeiler et al.~\cite{visualization13} and Krizhevsky et al.~\cite{alexnet} for more details. Figure~\ref{fig:activetemplatenet} shows the model used in our active template network. The difference from~\cite{visualization13} is the loss function we use. Instead of a classification loss, we predict the normalized coefficients by minimizing $\ell_2$ distance between the prediction and the ground truth coefficients. Suppose the predicted coefficients are denoted as $\bar{\boldsymbol{\alpha}}_{i,k}$ and the ground truth coefficients as $\hat{\boldsymbol{\alpha}}_{i,k}$. The $\ell_2$ loss  is defined as
\vspace{-3mm}
\begin{equation}
    J = \frac{1}{n}\sum_{i=1}^{n}\sum_{k=1}^{K}||\hat{\boldsymbol{\alpha}}_{i,k} - \bar{\boldsymbol{\alpha}}_{i,k}||^2.
\end{equation}
\vspace{-3mm}

 \begin{figure*}[t]
 \begin{center}
 \includegraphics[scale=0.6]{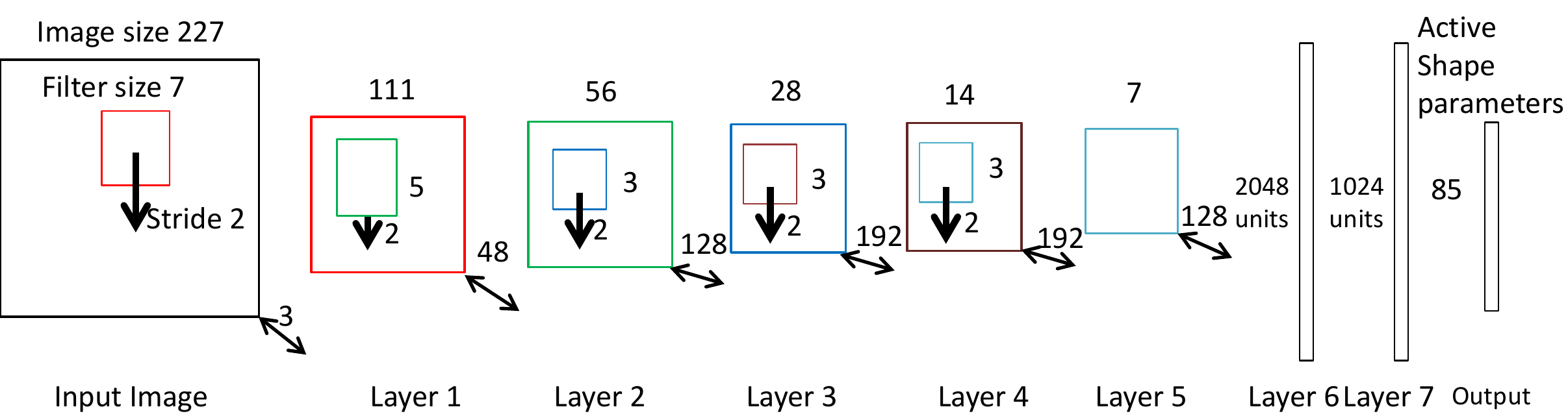}
 \vspace{-3mm}
 \caption{{Architecture of our active shape network. We take a $227\times 227\times3$ image as the input and convolve it with $48$ different 1st layer filters (red), each of which with the size $7\times 7$, using a stride of 2 in both dimensions. The obtained feature maps are then passed through a rectified linear function (not shown) to get $48$ different $111\times 111$ feature maps. Similar operations are repeated in 2nd, 3rd, 4th, 5th layers. The last two layers are fully-connected with 2048 units and 1024 units, respectively. The output layer with $85 = 17\times 5$ units is a regression function with $\ell_2$-norm for $K = 17$ semantic labels and each with 5 dimensions, including positions, scales and visibility flag. }}
 \vspace{-8mm}
 \label{fig:activeshapenet}
 \end{center}
 \end{figure*}
 
The network parameters (filters in the convolutional layers, weight matrices in the fully-connected layers and biases) are trained by Back-propagation. For the simplicity, we eliminate the subscript $i$ for each image in the following. Given an input image $x$, our active template network can predict template coefficients $\{\bar{\boldsymbol{\alpha}}_{k}\}_1^k$ for all labels and then we obtain the absolute coefficients $\{\tilde{\boldsymbol{\alpha}}_{k}\}_1^k$ by using the inverse function of Eq.(\ref{eq:normalize}). The normalized mask $\boldsymbol{b}_{k}$ for each label can be reconstructed by the linear combination of the specific template dictionary with $\tilde{\boldsymbol{\alpha}}_{k}$, as $\boldsymbol{b}_{k} = D_k\tilde{\boldsymbol{\alpha}}_{k}$.
 
\vspace{-4.5mm}
\subsection{Active Shape Network}
\vspace{-1mm}

After obtaining the normalized mask of each label, we need to morph them into the more precise masks at accurate positions in the image. In this paper, we denote the positions, scales and visibilities of the label masks as the active shape parameters $\{\boldsymbol{s}_k\}_1^K$, predicted by our active shape network. The structure outputs $\boldsymbol{s}_k = (b^x_k, b^y_k, b^w_k, b^h_k, v_k)$ include the top-left coordinates $(b^x_k, b^y_k)\in \mathbb{R}^2$, the width $b^w_k$, the height $b^h_k$ and the visibility flag $v_k$ which is set as $1$ if the $k$-th label appears in the image.

Figure~\ref{fig:activeshapenet} shows the architecture of our active shape network. The first convolutional layer filters a $227\times227\times3$ input image with $48$ kernels of size $7\times 7\times 3$ with a stride of $2$ pixels. The second convolutional layer takes the rectified output of the first convolutional layer as the input and filters it with $128$ kernels of size $5\times5\times48$ with a stride of $2$ pixels. The 3rd, 4th and 5th convolutional layers are connected to one another, and the 3rd and 4th layers are also with a stride of $2$ pixels. The last two fully-connected layers have $2048$ and $1024$ units, respectively. The output layer predicts $\{s_k\}_1^K$ for all labels, resulting in $85$ units. Furthermore, since the positions and scales are in absolute coordinates, it will be beneficial to normalize them with respect to the mean and standard deviations of positions and scales, similar as in Eq.(\ref{eq:normalize}). We keep the original values of visibility flags which are either 1 or 0. We minimize $\ell_2$ distance between the prediction and the ground truth parameters. Suppose the predicted parameters are denoted as $\bar{s}_{k}$ and the ground truth parameters as $\hat{s}_{k}$. The corresponding $\ell_2$ loss is defined as
 
 \vspace{-4mm}
\begin{equation}
    J = \frac{1}{n}\sum_{i=1}^{n}\sum_{k=1}^{K}||\hat{\boldsymbol{s}}_{k} - \bar{\boldsymbol{s}}_{k}||^2.
\end{equation}
\vspace{-3mm}

The previous infrastructures for the classification tasks~\cite{alexnet}~\cite{visualization13} include the max-pooling layer to make the network invariant to scale/translation changes and reduce the scale of feature maps. However, our network for regressing shape parameters is sensitive to position variance. To remedy this problem, our network eliminates the pooling layer and keeps the same overall depth of the network with~\cite{visualization13}. The new architecture retains much more information in the first few layers (e.g. the feature map with size $111\times111$ vs $55\times55$  in the 1st layer and $56\times56$ vs $14\times14$  in the 2nd layer, compared with the model in Figure.~\ref{fig:activetemplatenet}). Additionally, we reduce the scale of feature maps gradually, using a stride of $2$ pixels as well in the 2nd, 3rd, and 4th layers. Given that our dataset is much smaller than the ImageNet dataset, we decrease the filter number in each convolution layer and the size of the fully-connected layers to prevent over-fitting.

The contextual interactions between all semantic label masks (e.g. label exclusiveness and spatial layouts) are intrinsically captured by all of the hidden layers.
Given a test image $x$, the active shape network predicts shape parameters $\{\bar{\boldsymbol{s}}_{k}\}_1^K$ for all label masks and the absolute image coordinates $\{\tilde{\boldsymbol{s}}_{k}\}_1^k$ are obtained by using the inverse normalization. 

\textbf{Bounding Box Refinement.} In addition, considering the prediction error of shape parameters, we utilize the bounding box refinement to further reduce the mislocalizations. Specifically, we train $K$ linear regression models to predict new positions (e.g., $b^x_k, b^y_k, b^w_k, b^h_k$) for all labels, following the method proposed for object detection~\cite{regionCNN14}. To train the bounding box regressor for each label, all the training images are cropped around the predicted positions and then enlarged by a factor of $1.5$ to contain more surrounding context. The input for training is a set of training pairs, i.e., the predicted positions from our network and the ground-truth bounding boxes for each label. Note that only the predicted label mask which has an over $0.5$ overlap ratio with the ground-truth box is considered. The features for each training image are extracted from the outputs of the fully-connected layer of the ImageNet model~\cite{alexnet}. Finally, we use the same strategy to learn the position transformation. Please refer to~\cite{regionCNN14} for more details.

\vspace{-4mm}
\subsection{Structure Output Combination and Super-pixel Smoothing}

After feeding the image into the above two networks, we can obtain the normalized mask $\boldsymbol{b}_k$ and the shape parameters $\boldsymbol{s}_k$ for each label. The confidence map $\boldsymbol{c}_k$ of each label $k$ is obtained by morphing the mask $\boldsymbol{b}_k$ into the absolute image coordinates with $\boldsymbol{s}_k$. Note that the visibility flag $v_k$ denotes whether the $k$-th label appears in the image or not. Only if the visibility flag satisfies $v_k \geq 0.5$, the associated masks are considered. Note that this threshold is only used to prune the less likely appeared label masks. The final label masks are mainly decided by the predicted template coefficients and active shape parameters.

\begin{figure}
\begin{center}
\includegraphics[scale=0.5]{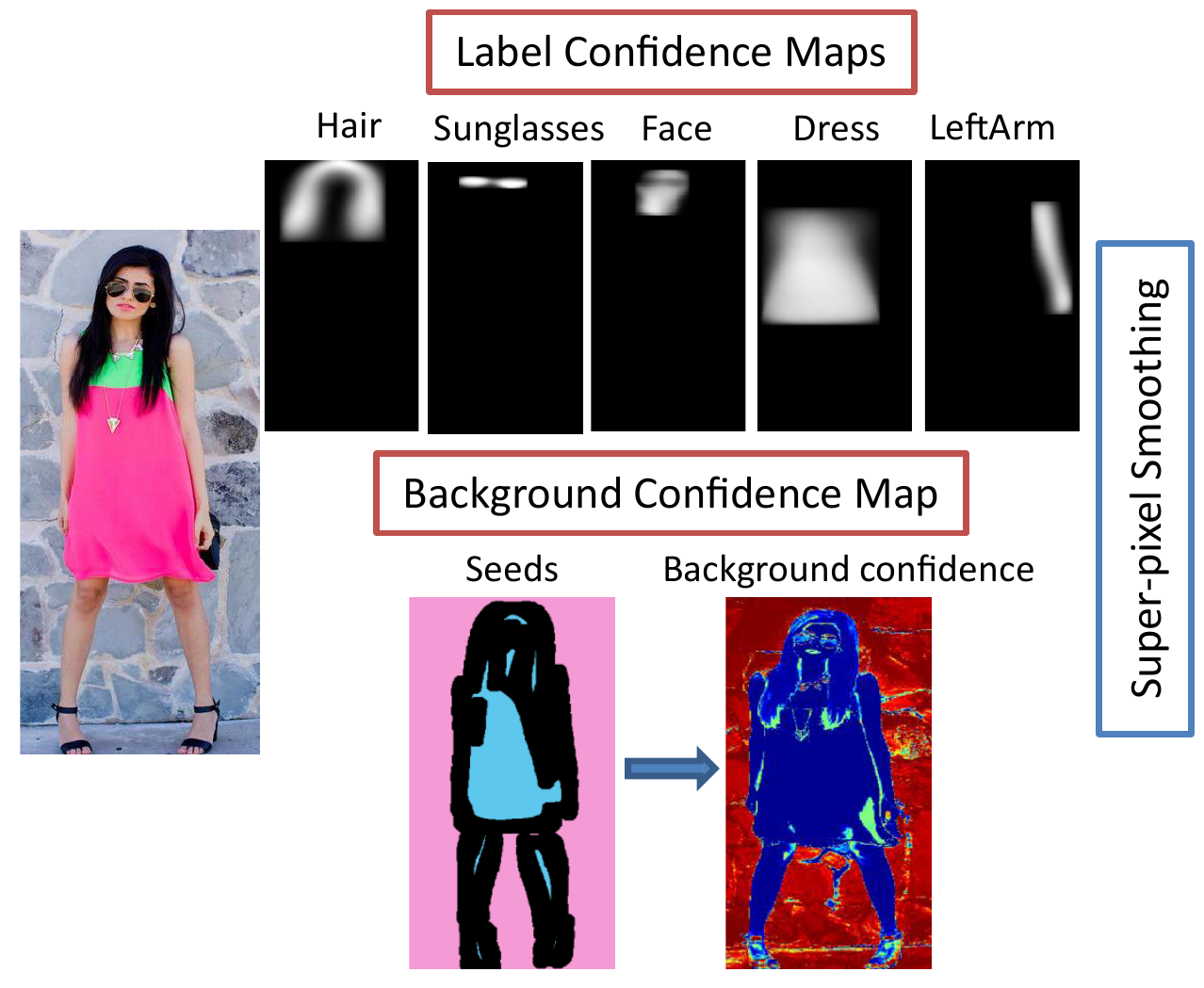}
\vspace{-3mm}
\caption{{Our structure output combination. The confidence maps of all foreground labels are predicted by fusing two types of structure outputs. Then we can produce the background confidence map: we first generate the foreground (blue pixels) and background seeds (pink pixels) and then predict the background confidence map (the red colored pixels have the highest probability for background). Finally the superpixel smoothing is used to refine the parsing result.}}
\vspace{-8mm}
\label{fig:combination}
\end{center}
\end{figure}

Our network can only predict the confidence maps of all foreground labels. For the background label, we predict its probability for each pixel by adopting the interactive image segmentation method~\cite{gscsegmentation10}. We automatically obtain the reliable foreground and background seeds from the confidence maps of all labels. Specifically, we first calculate the foreground confidence map $\boldsymbol{c}_f$ by maximizing the confidences of each label as $\boldsymbol{c}_f = \max_{k=1}^K \boldsymbol{c}_k$. Only the pixels of $\boldsymbol{c}_f$ with the confidence larger than $0.5$ is regarded as the foreground. Then the erode operation with the filter size $10$ based on the foreground mask is performed to produce the foreground seeds, displayed as the blue pixels of seed images in Figure~\ref{fig:combination}. The background seeds are obtained by dilating the inverse of the foreground mask within $10$ neighborhoods,  displayed as the pink pixels of seed images in Figure~\ref{fig:combination}. Based on the seeds, we can predict the background confidence map by learning the color model as in~\cite{gscsegmentation10}. 

\textbf{Super-pixel Smoothing.} To combine the confidence maps of all semantic labels and the background, we apply super-pixel smoothing and refine the parsing results for more precise pixel-level segmentation. In particular, our approach first computes an over-segmentation of the input image using a fast segmentation algorithm~\cite{FelzenszwalbH04}. We denote the background label as $k=0$ and thus we have $K+1$ possible labels for each pixel $i'$. The confidence map set is denoted as $\boldsymbol{C} = \{\boldsymbol{c}_k\}_{k=0}^{K}$, where $\boldsymbol{c}_0$ is the obtained background confidence map using~\cite{gscsegmentation10}. The super-pixel which contains the pixel~$i'$ is defined as $q_{i'}$ and the predicted label of the pixel~$i'$ is denoted as $y_{i'}$. Our final parsing result is thus calculated as
 
 \vspace{-3mm}
\begin{equation}
	y_{i'} = \max_k\sum_{j'\in q_{i'}}\boldsymbol{c}_k(j'),
\end{equation}
where $j'$ denotes each pixel in the super-pixel $q_{i'}$ and $\boldsymbol{c}_k(j')$ is the probability of the pixel $j'$ in the map $\boldsymbol{c}_k$. Since we only perform the maximization of the average confidences for all labels, our super-pixel smoothing method is very simple and fast.

\vspace{-5mm}
\section{Experiments} 
\label{sec:exp}

\begin{figure}
\begin{center}
\includegraphics[scale=0.4]{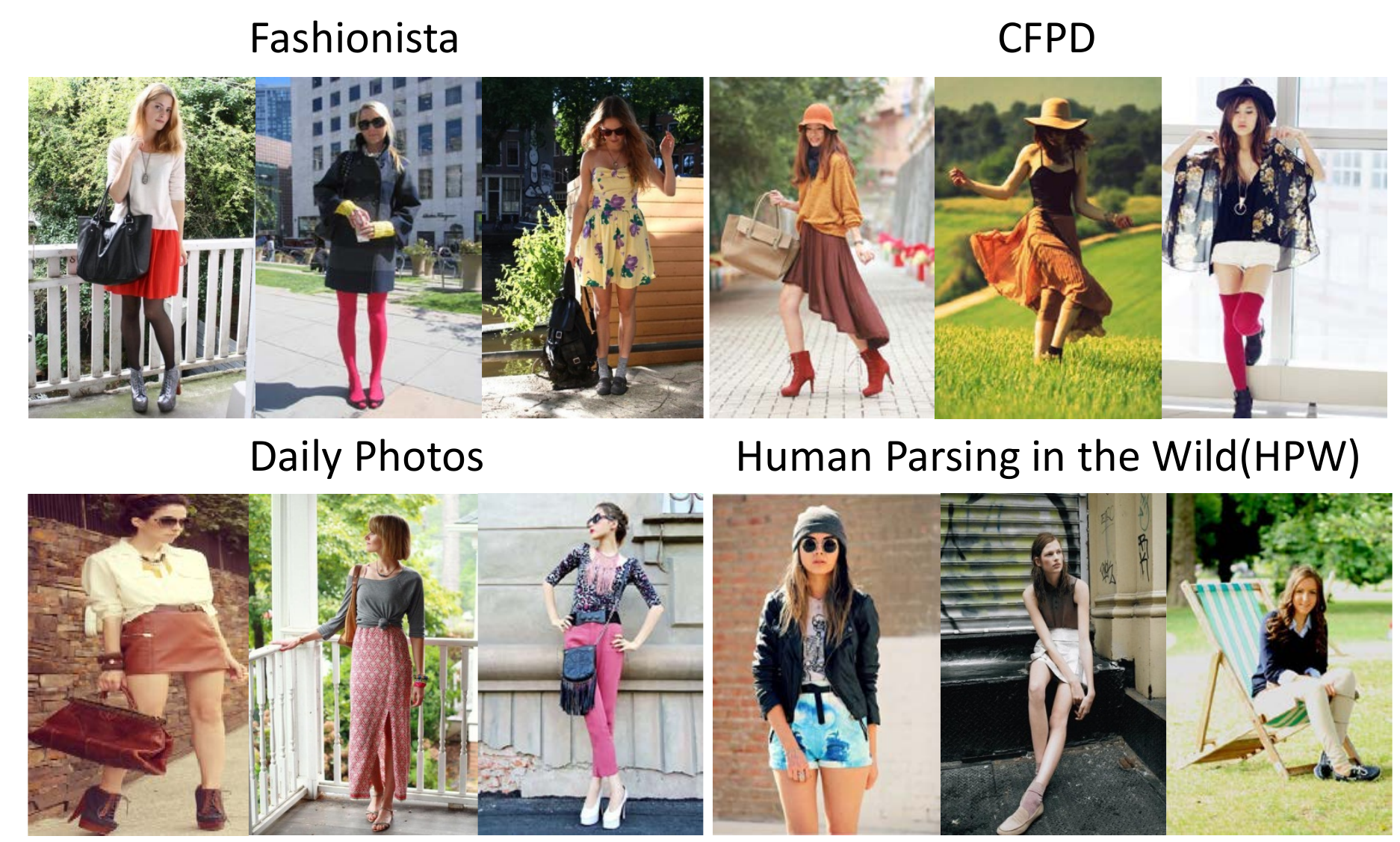}
\vspace{-2mm}
\caption{{Exemplar images in the combined dataset. 
}}
\vspace{-8mm}
\label{fig:dataset}
\end{center}
\end{figure}
\subsection{Experimental Settings}
\textbf{Datasets}: A large number of training samples are required for most of the deep models~\cite{alexnet}~\cite{regionCNN14}. However, existing public available datasets for human parsing are relatively small. The largest existing human parsing dataset, to our best knowledge, contains only $2,682$ images, which is insufficient for training a robust deep network model. Thus, we combine data from three small benchmark datasets: (1) the Fashionista dataset~\cite{yamaguchi2012parsing} containing 685 images, (2)~the Colorful Fashion Parsing Data (CFPD) dataset~\cite{siparsingTMM14} containing 2,682 images, and (3) the Daily Photos dataset~\cite{Dongparsing13} containing $2,500$ images. All images in these three datasets contain standing people in frontal/near-frontal view with good visibilities of all body parts. Following the label set defined by Dong et al.~\cite{Dongparsing13}, we merge the labels of Fashionista and CFPD datasets to $18$ categories: face, sunglass, hat, scarf, hair, upper-clothes, left-arm, right-arm, belt, pants, left-leg, right-leg, skirt, left-shoe, right-shoe, bag, dress and background. To enlarge the diversity of our dataset, we crawl another $1,833$ challenging images to construct the Human Parsing in the Wild (HPW) dataset and annotate pixel-level labels following ~\cite{Dongparsing13}. As shown in Figure~\ref{fig:dataset}, our newly annotated data are mainly more realistic images containing challenging poses (e.g. sitting) and occlusion, which is a good supplement to the existing three datasets. The final combined dataset from the four datasets contains $7,700$ images. We use $6,000$ images for training, $1,000$ for testing and $700$ as the validation set. The occurrences of each label in our collected dataset are reported in Table~\ref{tab:F1scores}. For fair comparison with published algorithms, we use the same evaluation criterion as in~\cite{Yamaguchiparsing13}, which contains accuracy, average precision, average recall, and average F-1 scores over pixels.

\begin{table*}[htbp]\setlength{\tabcolsep}{1pt}
  \centering
 \caption{Comparison of parsing performances with several architectural variants of our model and two state-of-the-arts. }\label{tab:tableoverall} \label{table:comparison}
 \vspace{-3mm}
    \begin{tabular}{cccccccccccccccccccccc}
    \toprule
          \textbf{Method} &  \textbf{Accuracy}   &  \textbf{F.g. accuracy}  &  \textbf{Avg. precision}   &    \textbf{Avg. recall}  &  \textbf{Avg. F-1 score} \\
    \midrule
    Yamaguchi et al.~\cite{yamaguchi2012parsing} (456) & 82.54 & 46.70 & 31.67 & 43.74 & 35.78\\
    PaperDoll~\cite{Yamaguchiparsing13} (456) & 86.74 & 50.34 & 43.38 & 41.21 & 37.54\\     
    Yamaguchi et al.~\cite{yamaguchi2012parsing} (6000) & 84.38 & 55.59 & 37.54 & 51.05 & 41.80 \\
    PaperDoll~\cite{Yamaguchiparsing13} (6000) & 88.96 & 62.18 & 52.75 & 49.43 & 44.76 \\
    Yamaguchi et al.~\cite{yamaguchi2012parsing} (6000 test 229) & 87.87 & 58.85 & 51.04 & 48.05 & 42.87 \\
    PaperDoll~\cite{Yamaguchiparsing13} (6000 test 229) & 89.98 & 65.66 & 54.87 & 51.16 & 46.80 \\
    \midrule
    ATR (unified) & 84.95 & 45.65 & 51.90 & 33.07 & 38.62\\
    \midrule
    ATR (PCA) & 86.43 & 52.83 & 63.50 & 43.39 & 48.87\\
    {ATR (NMF$\ell_1$)}& 88.49 & 61.44 & 62.00 & 49.64 & 53.77\\
    \midrule
      ATR (zeilernet) & 88.59 & 60.77 & 62.66 & 48.55 & 53.62\\
      ATR (lessfc) & 90.16 & 67.74 & 68.17 & 56.59 & 60.50\\ 
      ATR (lessfcfilters)& 90.21
              & 67.17 & 69.16 & 56.04 & 60.77\\
    {ATR (nopool)} & 91.01 & 70.40 & 69.61 & 58.82 & 62.78\\
    \midrule
    \textbf{ATR (noSPR)} & 89.33 & 64.79 & 63.75 & 56.19 & 59.60\\
    \textbf{ATR} & \textbf{91.11} & \textbf{71.04} & \textbf{71.69}
      & \textbf{60.25} & \textbf{64.38}\\
    {ATR (test 229)} & {92.33} & {76.54} & {73.93} & {66.49} & {69.30}\\  
    \midrule
     Upperbound  &  98.67 & 93.61 & 95.45 & 92.79 & 94.04\\
    \bottomrule
    \end{tabular}%
  \vspace{-3mm}
\end{table*}%

\begin{table*} [htbp]\setlength{\tabcolsep}{2pt}
\centering
{
\caption{F-1 scores of foreground semantic labels. Comparison of F1-scores with several architectural variants of our model and two state-of-the-art methods.}\label{tab:F1scores}
\vspace{-3mm}
\begin{tabular}{cccccccccccccccccccccc}
  \toprule
  Method & Hair & Bag & Belt & Dress & Face & Hat & L-arm & L-leg & L-shoe & Pants & R-arm & R-leg & R-shoe & Scarf & Skirt & S-gls & U-cloth\\
 \midrule
  Label occurrences & 7059 & 3517 & 1952 & 2303 & 7387 & 1918 & 6956 & 5330 & 6146 & 3501 & 6615 & 5571 & 6203 & 440 & 2484 & 2221 & 5933\\
 \midrule
  Yamaguchi et al.~\cite{yamaguchi2012parsing}\\ (6000) & 59.96 & 24.53 & 14.68 & 40.94 & 72.10 & 8.44 & 45.33 & 58.52 & 38.24 & 55.42 & 46.65 & 57.03 & 38.33 & 11.43 & 17.57 & 12.09 & 56.07\\
  PaperDoll~\cite{Yamaguchiparsing13}\\ (6000) & 63.58 & 30.52 & 16.94 & 59.49 & 61.63 & 1.72 & 45.23 & 52.19 & 45.79 & 69.35 & 46.75 & 55.60 & 44.47 & 2.95 & 40.20 & 0.23 & 71.87\\
  Yamaguchi et al.~\cite{yamaguchi2012parsing}\\ (6000 test 229) & 62.58 & 27.31 & 18.50 & 54.26 & 60.26 & 1.48 & 42.96 & 47.93 & 44.83 & 66.37 & 45.17 & 52.22 & 44.01 & 2.44 & 35.49 & 0.19 & 68.98\\
  PaperDoll~\cite{Yamaguchiparsing13}\\ (6000 test 229) & 64.45 & 31.22 & 16.78 & 65.42 & 62.32 & 2.12 & 48.20 & 56.16 & 46.79 & 73.51 & 48.62 & 58.35 & 45.40 & 3.93 & 47.17 & 0.28 & 74.36\\
 \midrule
   {ATR (unified)} & 58.87 & 4.47 & 11.74 & 59.25 & 63.74 & 34.10 & 30.64 & 45.96 & 18.28 & 46.60 & 31.27 & 50.69 & 20.08 & 19.25 & 35.62 & 1.79 & 69.33\\
 \midrule
  {ATR (PCA)} & 59.12 & 45.53 & 5.27 & 53.74 & 65.42 & 42.06 & 51.60 & 64.04 & 47.47 & 60.84 & 49.76 & 60.59 & 43.88 & 17.63 & 43.29 & 5.37 & 69.73
 \\
  {ATR (NMF$\ell_1$)} & 56.04 & 42.06 & 5.70 & 74.90 & 63.74 & 65.47 & 50.84 & 62.90 & 47.16 & 70.26 & 43.42 & 61.96 & 46.62 & 28.57 & 73.34 & 7.58 & 72.28\\
 \midrule
  {ATR (zeilernet)} & 63.78 & 31.33 & 1.13 & 73.43 & 69.02 & 63.82 & 45.89 & 60.86 & 41.83 & 70.18 & 42.74 & 65.68 & 38.84 & 46.89 & 66.02 & 13.37 & 74.87\\
 
  {ATR (lessfc)} & 67.55 & 39.42 & 17.79 & 77.85 & 72.28 & 71.26 & 51.13 & 63.90 & 52.09 & 77.82 & 51.75 & 69.12 & 44.63 & 60.58 & 79.13 & 18.44 & 78.02\\
 
  {ATR (lessfcfilters)}  & 67.54 & 36.93 & 21.80 & 78.10 & 72.12 & 73.26 & 57.23 & 66.43 & 50.73 & 76.39 & 55.44 & 67.30 & 48.74 & 47.29 & 77.83 & 22.66 & 77.58\\
  {ATR (nopool)} & \textbf{71.67} & \textbf{56.59} & 14.31 & \textbf{82.15} & \textbf{76.53} & 59.18 & \textbf{57.41} & \textbf{69.36} & 47.73 & 77.94 & \textbf{60.73} & 69.98 & 48.72 & 53.16 & 79.89 & 33.69 & \textbf{79.50}\\
  \midrule
  \textbf{ATR (noSPR)} & 69.11 & 49.79 & 18.00 & 76.63 & 74.55 & 68.61 & 49.17 & 59.95 & 47.21 & 72.29 & 52.07 & 63.04 & 45.87 & 45.85 & 73.87 & \textbf{35.66} & 75.21\\
 
  \textbf{ATR} & 68.18 & 53.66 & \textbf{22.88} & 82.02 & 74.71 & \textbf{77.97} & 53.79 & 69.07 & \textbf{53.51} & \textbf{79.77} & 58.57 & \textbf{71.69} & \textbf{50.26} & \textbf{57.07} & \textbf{80.36} & 29.20 & 79.39\\
  {ATR (test 229)} & 69.35 & 66.91 & 30.50 & 85.38 & 78.48 & 77.14 & 64.37 & 74.56 & 57.76 & 82.96 & 63.25 & 76.07 & 55.87 & 63.26 & 83.35 & 38.14 & 82.77\\
 \bottomrule
\end{tabular}
\vspace{-7mm}
}
\end{table*}

\textbf{Data Augmentation}: To reduce over-fitting on image data, we manually enlarge the training data to increase the diversity using the translations and horizontal reflections. Specifically, we first detect the bounding box of the human body~\cite{regionCNN14} and then incrementally cover more context outside the box with the stride of $20$ pixels in eight directions (i.e. top/down, left/right, topleft/topright, downleft/downright). In addition, we enlarge the scale of the detected bounding box with three factors, i.e., $1.2$, $1.5$, $1.8$. The horizontal reflections are used for all the cropped images. Then we resize all these images into $227\times 227\times3$ using the nearest-neighbor interpolation. This increases size of our
training set by a factor of $24 = (8+3)\times 2$. Although the resulting training examples are highly inter-dependent, the data augmentation can significantly increase the diversity of features, especially for predicting the active shape parameters. 

\textbf{Implementation Details}: Our two networks aim to predict the masks and shape parameters of $K = 17$ labels. To learn the template dictionary for each label, we normalize the binary mask into a regularized size $r^w$ and $r^h$ as $100$ and the template number $M$ as $50$. The penalty $\lambda$ for the NMF is set as $0.001$. When the training image does not have certain labels, we set their corresponding template coefficients and shape parameters as zeros. We implement the two networks under the caffe framework~\cite{caffe13} and train them using stochastic gradient descent with a batch size of $128$ examples, momentum of $0.9$, and
weight decay of $0.0005$. We use an equal learning rate for all layers and adjust it manually. The strategy is to divide the learning rate by $10$ when the validation error rate stops decreasing with the current learning rate. The learning rate is initialized at $0.0005$ for the two networks. We train the networks for roughly $120$ epochs, which takes $2$ to $3$ days on one NVIDIA GTX TITAN 6GB GPU. Our algorithm can rapidly process one $227\times227$ image within about $0.5$ second, as measured on a NVIDIA GTX TITAN 6GB GPU. This compares favorably to other approaches, as some of the current state-of-the-art approaches
have higher complexity: ~\cite{yamaguchi2012parsing} runs in about $10$ to $15$ seconds,
while ~\cite{Dongparsing13} runs in $1$ to $2$ minutes. 

\vspace{-5mm}
\subsection{Results and Comparisons}
\vspace{-1mm}

 We compare our ATR framework with the two state-of-the-art works~\cite{Yamaguchiparsing13}~\cite{yamaguchi2012parsing}. We use their public available codes and carefully tune the parameters according to~\cite{yamaguchi2012parsing}~\cite{Yamaguchiparsing13} and train their models with the same training images as our method for the fair comparison. Note that Dong et al.~\cite{Dongparsing13} is not compared in this work because our experiments show the PaperDoll~\cite{Yamaguchiparsing13} can achieve the accuracy of $87.8\%$ on the 229 test images of the Fashionista dataset, which is better than the accuracy of $86\%$ reported in ~\cite{Dongparsing13} with the same label set. We implement two versions of our method. (1) ``ATR (noSPR)": the parsing results are obtained by maximizing the all confidence maps where no Super-Pixel Refinement (SPR) is used. (2) ``ATR": we refine the parsing results with the super-pixel smoothing. The results are listed in Table~\ref{tab:tableoverall}. 

The method of Yamaguchi et al.~\cite{yamaguchi2012parsing} and the PaperDoll~\cite{Yamaguchiparsing13} with 456 training images as on the public Fashionista dataset achieve $35.78\%$ and $37.54\%$ of average F1-score on evaluating our 1,000 test images, respectively. When training the model with more data (e.g., 6,000 images), the performances of the two baselines can be increased by 6.02$\%$~\cite{yamaguchi2012parsing} and 7.22$\%$~\cite{Yamaguchiparsing13}. However, our ``ATR" can significantly outperform these two baselines by over $22.58\%$ for Yamaguchi et al.~\cite{yamaguchi2012parsing} and $19.62\%$ for PaperDoll~\cite{Yamaguchiparsing13}. Our method also gives a huge boost in foreground accuracy: the two baselines achieve $55.59\%$ for Yamaguchi et al.~\cite{yamaguchi2012parsing} and $62.18\%$ for PaperDoll~\cite{Yamaguchiparsing13} while ``ATR" obtains $71.04\%$. ``ATR" also obtains much higher precision (71.69\% vs 37.54\% for~\cite{yamaguchi2012parsing} and 52.75\% for~\cite{Yamaguchiparsing13}) as well as higher recall (60.25\% vs 51.05\% for~\cite{yamaguchi2012parsing} and 49.43\% for~\cite{Yamaguchiparsing13}). The pixel-level accuracy is also increased by at least $2.15\%$. This verifies the effectiveness of our algorithm though it does not require explicit definition of any contextual relations and incorporation of complicated prior knowledge. For ``ATR (noSPR)", it also achieves superior performance than the baselines. The superiority of ``ATR (noSPR)" over the baselines demonstrates that our network has the capability of directly predicting reasonable label masks without any low-level segmentation methods which are commonly used by all previous methods. The improvements from ``ATR (noSPR)" to ``ATR" show that the super-pixel smoothing enables the parsing result to preserve more accurate boundary information. For the fair comparison, we also report the parsing results on the 229 test images of the Fashionista dataset~\cite{yamaguchi2012parsing}. Our method ``ATR (test 229)"  can also significantly outperform these two baselines by over $26.43\%$ for ``Yamaguchi et al.~\cite{yamaguchi2012parsing} (6000 test 229)" and $22.5\%$ for PaperDoll~\cite{Yamaguchiparsing13} (6000 test 229)" of average F1-score on evaluating 229 test images. This speaks well that our collected dataset contains much more realistic images with the challenging poses and occlusions than the small Fashionista dataset~\cite{yamaguchi2012parsing}.  

\begin{figure*}[t!]
\begin{center}
\includegraphics[scale=1.25]{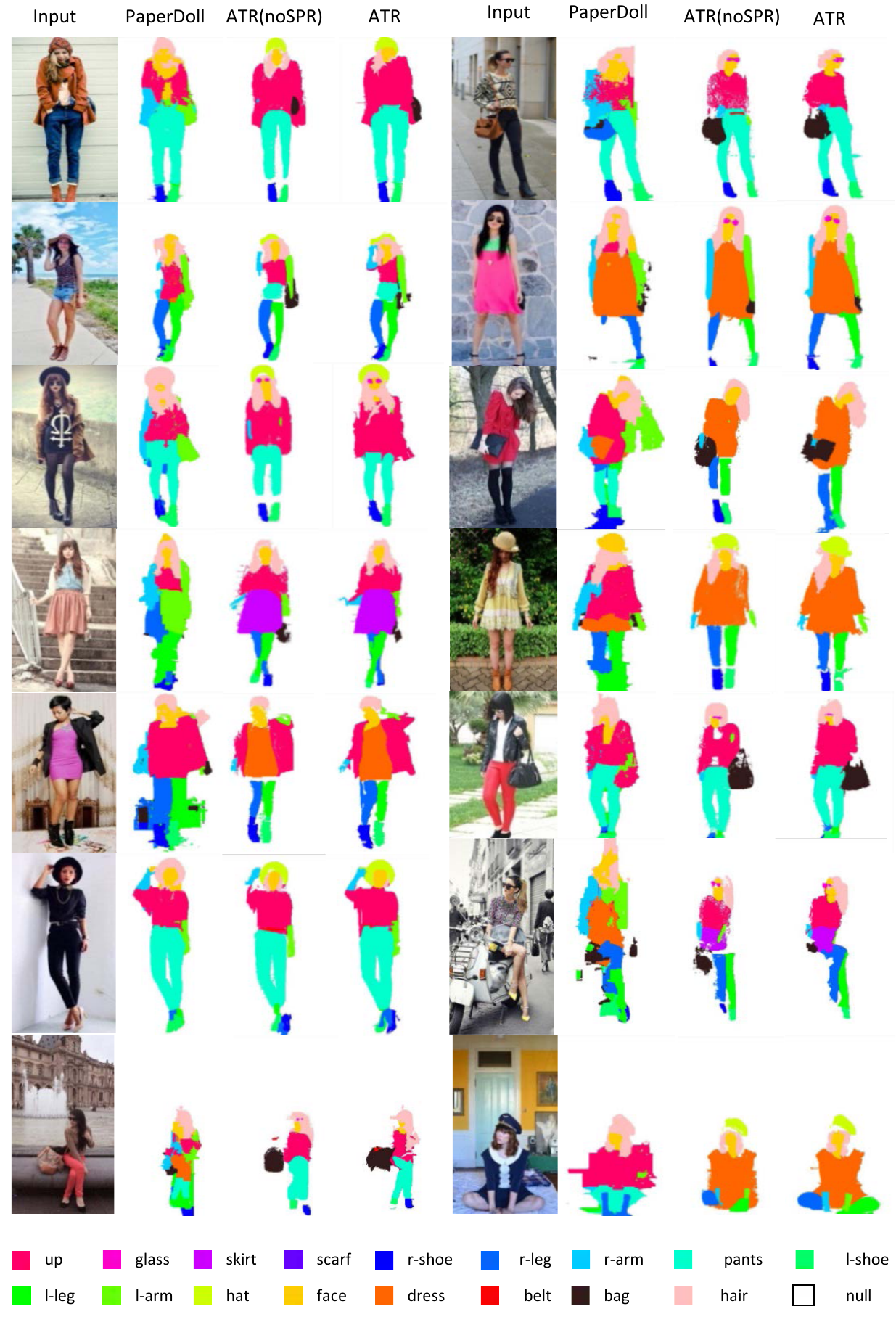}
\caption{{Comparison of parsing results with the state-of-the-art method and our two versions. For each image, we show the parsing results by PaperDoll~\cite{Yamaguchiparsing13}, our ``ATR (noSPR)" with no super-pixel smoothing and our full method ``ATR"  sequentially.}}
\label{fig:results}
\end{center}
\vspace{-0.4in}
\end{figure*}

We also present the F1-scores for each label in Table~\ref{tab:F1scores}. Generally, both versions of our method show much higher performance than the baselines. In terms of predicting small labels such as hat, belt, bag and scarf, our method achieves a large gain, e.g. 57.07\% vs 11.43\%~\cite{yamaguchi2012parsing} and 2.95\%~\cite{Yamaguchiparsing13} for scarf, 53.66\% vs 24.53\%~\cite{yamaguchi2012parsing} and 30.52\%~\cite{Yamaguchiparsing13} for bag. It demonstrates that our two networks can capture the internal relations between the labels and robustly predict the label masks with various clothing styles and poses. The qualitative comparison of parsing results is visualized in Figure~\ref{fig:results}. Our methods predict much more reasonable and meaningful label masks than the PaperDoll method~\cite{Yamaguchiparsing13} despite their large appearance and position variations. We can successfully predict small labels (e.g. sun-glasses, hat) while the PaperDoll~\cite{Yamaguchiparsing13} often fails and confuses them with the neighboring regions. For example, for the left image of the third row in Figure~\ref{fig:results}, we can detect sunglasses and hat while the PaperDoll totally misses them. The parsing results of our methods are cleaner and label masks bears strong semantic meanings while the results of~\cite{Yamaguchiparsing13} are heavily influenced by the low-level information, such as image clarity and color similarity. It demonstrates that our framework performs better in solving the high-level human parsing problem than the models based on low-level features. Finally, comparing the results of ``ATR (noSPR)" and ``ATR", we can find that ``ATR" can provide refined parsing results with respect to the region boundary. For example, for the left image in the first row in Figure~\ref{fig:results}, ``ATR" with super-pixel smoothing can effectively fill the gaps between the shoes and pants. 

\begin{table*} [htbp]\setlength{\tabcolsep}{0.4pt}
\centering\scriptsize{\caption{Detailed experimental settings by varying the model architectures of our networks.}\label{tab:shapesettings}
\vspace{-3mm}
\begin{tabular}{ c|c |c c |c c  c c | cc }
  \toprule
  \multicolumn{1}{c}{}  & \multicolumn{1}{c}{} & \multicolumn{2}{c}{Active Template} & \multicolumn{4}{c}{Active Shape} & \multicolumn{2}{c}{\tabincell{c}{Structure Output\\Combination}}\\
   \cmidrule(l){3-10}
    & ATR (unified) & ATR (PCA) & ATR (NMF$\ell_1$)& ATR (zeilernet) & ATR (lessfc) & ATR (lessfcfilters) & ATR (nopool) & ATR (noSPR) & ATR \\
  \hline
  Template generation & NMF & PCA & \tabincell{c}{NMF\\with $\ell_1$-norm} & NMF & NMF & NMF & NMF & NMF & NMF\\

  AT net & No & ours & ours & ours & ours & ours & ours & ours & ours\\

   AT output num & No & 850 & 850 & 850 & 850 & 850 & 850 & 850 & 850\\
 \hline
  AS net & NA & ours+BB & ours+BB & \tabincell{c}{Our replication\\ of~\cite{visualization13}} & \tabincell{c}{Adjust layers 6,7:\\2048,1024 units\\(based on~\cite{visualization13})} & \tabincell{c}{Adjust layers 1-5: \\48,128,192,192,128 \\ maps and layers 6,7:\\ 2048,1024 units\\(based on ~\cite{visualization13})} & ours & ours+BB & ours+BB\\
 
  AS output num& NA & 85 & 85 & 85 & 85 & 85 & 85 & 85 & 85\\
 \hline
  Unified network & \tabincell{c}{Our replication\\ of ~\cite{visualization13}} & NA & NA & NA & NA & NA & NA & NA & NA\\

  \tabincell{c}{Unified output num} & 935 & NA & NA & NA & NA & NA & NA & NA & NA\\
 \hline
  \tabincell{c}{Structure Output\\ Combination} & \tabincell{c}{SPR} & \tabincell{c}{SPR} & \tabincell{c}{SPR} & SPR & SPR & SPR & \tabincell{c}{SPR} & \tabincell{c}{MAP} & \tabincell{c}{SPR} \\
 \bottomrule
\end{tabular}
\vspace{-7mm}
}
\end{table*}

\vspace{-3mm}
\subsection{Ablation Studies of Our Networks}

We further evaluate the effectiveness of our two components of ATR, including the active template network and the active shape network, respectively.

\textbf{Active Template Network}: To justify the rationality of using the template coefficients rather than the binary label masks, we test the reconstruction errors in dictionary learning, named as ``Upperbound". The label masks are reconstructed using the ground truth template coefficients and the learned dictionaries, and all active shape parameters are fixed. Table~\ref{tab:tableoverall} shows that our ``Upperbound" can achieve $98.67\%$ in accuracy and $95.45\%$ in average precision. This well demonstrates that the strategy of representing the binary masks with the corresponding coefficients results in very few reconstruction errors. 

\begin{figure*}[t!]
\begin{center}
\includegraphics[scale=0.45]{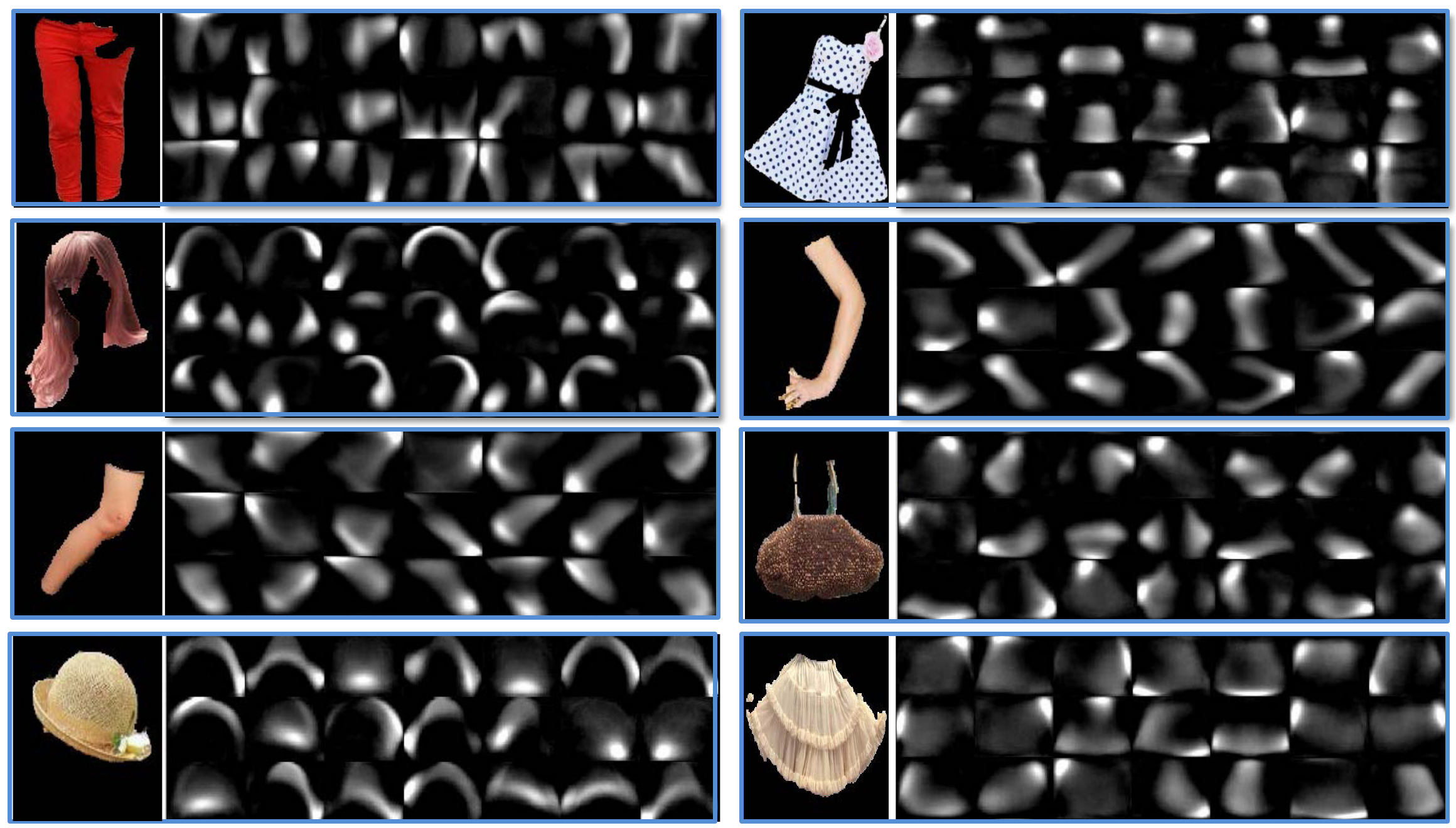}
\vspace{-3mm}
\caption{{Visualization of our template dictionaries of eight semantic labels, including pants, dress, hair, left-arm, right-leg, bag, hat and dress that are displayed sequentially. For each label, we display $21$ learned templates by the NMF method. Brighter pixels represent the most important parts for distinguishing different label masks. 
}}
\label{fig:maskcombination}
\end{center}
\vspace{-9mm}
\end{figure*}

We also evaluate other mask reconstruction approaches: ``ATR (PCA)" and ``ATR (NMF$\ell_1$)". The results are listed in Table~\ref{tab:tableoverall} and Table~\ref{tab:F1scores}. Table~\ref{tab:shapesettings} shows the details of experimental settings. First, we use the Principal Component Analysis (PCA) method~\cite{pca} for dictionary learning instead of the NMF, named as "ATR (PCA)". The same number of bases (i.e. 50 for each label) as in NMF is selected to construct the template dictionary. "ATR (PCA)" results in accuracy decrease by $4.68\%$ as well as $15.51\%$ in average F1-score, compared with ``ATR". PCA can be viewed as the eigenvector-based multivariate analysis that projects the data using only a few principle components, and the reconstruction coefficients and the basis vectors are either negative or positive. However, the NMF can learn the part-based decompositions and only additive combinations of templates are allowed, which is beneficial for our reconstruction. We also visualize our learned templates of each label as Figure~\ref{fig:maskcombination} shows. Most of the learned templates are in good shapes and bear strong semantic meanings. In addition, the templates are very diverse that can capture the large variances of label masks. These results verify that the nonnegative basis vectors can generate more expressiveness in the reconstruction. Second, to evaluate the effects of different norms upon the template coefficient prediction in Eq.(\ref{eq:nmf}), we use the $\ell_1$-norm for ``ATR (NMF$\ell_1$)" to yield more sparse template coefficients. Even though the $\ell_1$-norm has shown promising results in image reconstruction~\cite{peng2012rasl} and is commonly used in a wide range of computer vision problems, its performance is inferior to the ``ATR (SPR)" that uses the $\ell_2$-norm to constrain too many sparse values, that is, 88.49\% vs 91.11\% in accuracy. The possible reason may be that our network can hardly predict optimal values with the sparse coefficients which contain too many zeros.

\begin{figure*}[t!]
\begin{center}
\includegraphics[scale=0.5]{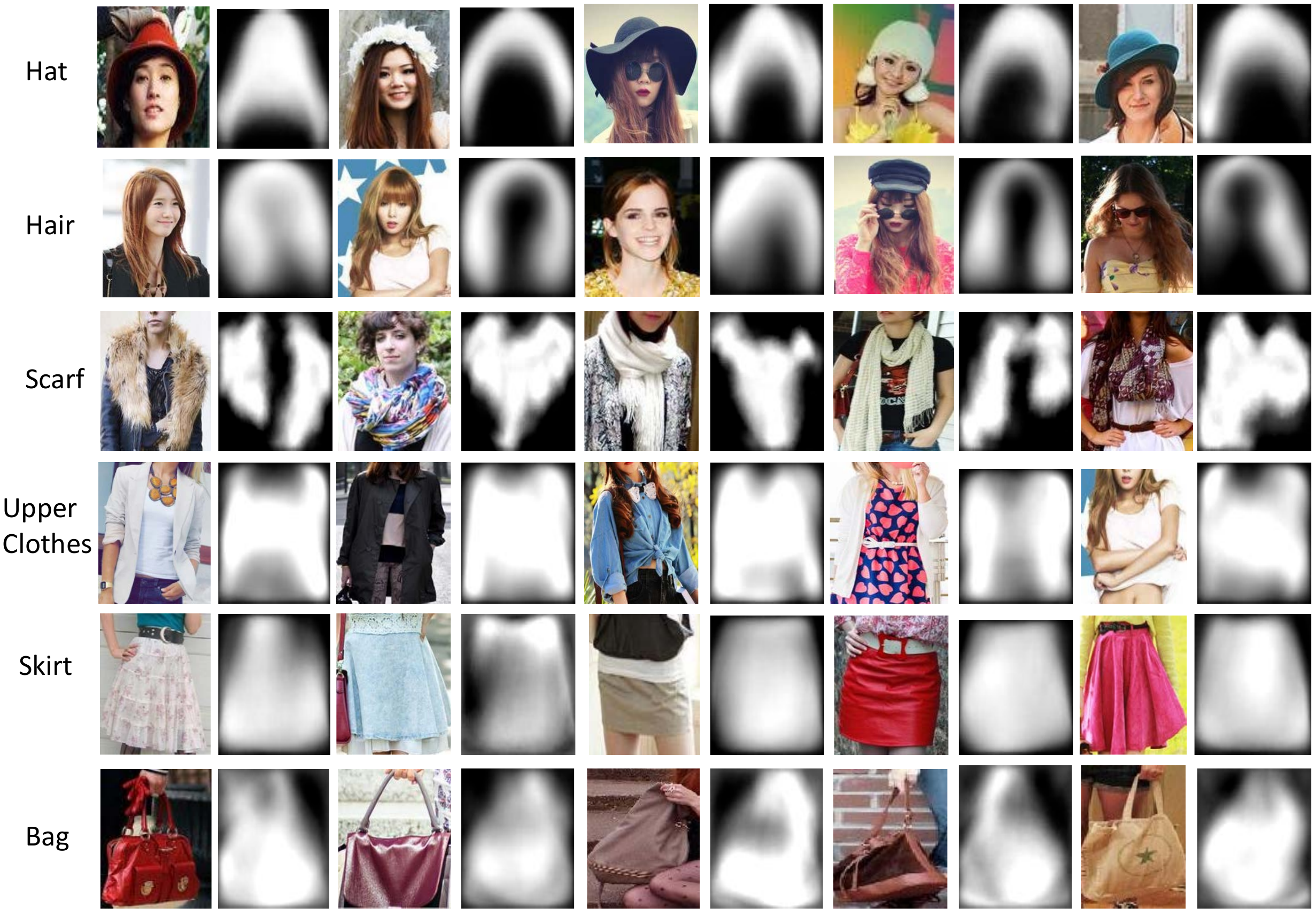}
\vspace{-2mm}
\caption{{Visualization of our predicted label masks with the active template network. We take the six semantic labels as the examples, such as hat, hair, scarf, upper-clothes, skirt and bag. The pixel with brighter color indicates that it is more likely to be assigned as the specific label. 
}}
\label{fig:activeshape}
\end{center}
\vspace{-0.3in}
\end{figure*}

Figure~\ref{fig:activeshape} visualizes the predicted label masks for $6$ semantic labels with our active template network. The pixel in each mask with brighter color indicates its larger probability to be the specific label. Our network performs well in predicting the various shapes of the label masks. In particular, the predicted masks of "hat" and "hair" are highly consistent with the ground truth masks, and the fine-grained shapes for each label can also be visually distinguished (e.g. long hair vs short hair). For example, the third row in Figure~\ref{fig:activeshape} shows several scarfs of different shapes. Even though the first scarf contains two disconnected regions and the second scarf is an entire region, our network can actively predict their respective shapes.

\textbf{Active Shape Network}: In Table~\ref{tab:tableoverall} and Table~\ref{tab:F1scores}, we also explore other model architectures for regressing the active shape parameters by adjusting the layer size gradually. We evaluate four cases of architectures: 1)~``ATR (zeilernet)" which follows the model architecture in~\cite{visualization13}; 2) ``ATR (lessfc)" where the size of two fully-connected layers is changed into 2048 and 1024 from the original 4096, respectively; 3) ``ATR (lessfcfilters)" where the number of filter maps is decreased by half, and also the size of fully-connected layers is changed as in ``ATR (lessfc)"; 4) ``ATR (nopool)" where the max-pooling layer is eliminated and the feature map size is gradually reduced by using the stride in the convolution layers, i.e., our proposed active shape network.  The performances of these settings are evaluated without the bounding-box refinement. In ``ATR (lessfc)" and ``ATR (lessfcfilters)", the model is trained from the scratch with the architecture in~\cite{visualization13}. Please refer to Table~\ref{tab:shapesettings} for more details of experimental settings. The ``ATR (zeilernet)" which uses the well-performed model infrastructure in image classification~\cite{visualization13} gives inferior performance to our network ``ATR (nopool)" (88.59\% vs 91.01\% in accuracy and 53.62\% vs 62.78\% in average F1-score). The main reason may be that the model for classification is not optimal for predicting our shape parameters which are sensitive to position variances. Besides, our dataset is much smaller than the ImageNet dataset. Using large layer size may result in over-fitting for our model. Thus we decrease the size of fully-connected layers, since they contain the majority of model parameters. The resulting accuracy and average F1-score of ``ATR (lessfc)" show significant increase by 1.57\% and 6.88\%, respectively, compared to ``ATR (zeilernet)". The ``ATR (lessfcfilters)" which decreases the number of filter maps yields slight performance improvements, but largely decreases the training parameter number. This suggests that a small number of filter maps is enough for training our model. Based on the ``ATR (lessfcfilters)", our final network ``ATR (nopool)" eliminates the max-pooling operation, such that more information is reserved in the first few layers. ``ATR (nopool)" gives a large gain in performance compared with ``ATR (lessfcfilters)" (91.01\% vs 90.21\% in accuracy and 62.78\% vs 60.77\% in average F1-score). This verifies the effectiveness of eliminating max-pooling layers for solving the position sensitive problems. Moreover, we test the effectiveness of the bounding box regression for obtaining better shape parameters, by comparing the results of ``ATR (nopool)" and "ATR". It shows that the bounding box refinement improves the average F1-score of ``ATR (nopool)" by $1.6\%$ by using fine-tuned active shape parameters of semantic labels.

\textbf{Discussion:} We evaluate the performance of training one unified network for regressing the template coefficients and active shape parameters. The ``ATR (unified)" version follows the network infrastructure in~\cite{visualization13} and targets on predicting all the structure outputs together. More details are presented in Table~\ref{tab:shapesettings}. The reported results in Table~\ref{tab:F1scores} are much worse than all other versions, especially than ``ATR" (84.95\% vs 91.11\% in accuracy and 38.62\% vs 64.38\% in average F1-score). The reason for the inferiority of the unified network may be that the learning of template coefficients and active shape parameters can be treated as two different tasks and often require different network architectures, as we design. The first task with max-pooling is essentially selecting the most appropriate templates for reconstructing label masks with the template dictionaries and the second one without max-pooling aims at predicting the precise locations. Particularly, our framework with two separated networks has shown significant improvement on performance than previous work~\cite{Yamaguchiparsing13} (increasing by 19.62$\%$ of F1-scores). The network for regressing active template coefficients and shape parameters together may further improve the performance by incorporating the complicated contextual interactions of label masks and their spatial layouts. But our experiment shows that directly combining two kinds of structure outputs works do not work well for human parsing. In the further works, we will explore how to design a more effective network architecture to combine these two networks.

\vspace{-5mm}
\section{Conclusions} \label{sec:conclusion}
\vspace{-2mm}
In this work, we formulate the human parsing task as an Active Template Regression problem. Two separate convolutional neural networks, namely, active template network and active shape network, are designed to build the end-to-end relation between the input image and the structure outputs. The first CNN network is with max-pooling to predict the mask template coefficients, while the second CNN network is without max-pooling for position sensitiveness to predict the active shape parameters. Extensive experimental results clearly demonstrated the effectiveness of the proposed ATR framework. In the future, we plan to further explore how to adequately utilize the low-level information (e.g. edges and super-pixels). In addition, we will integrate the fine-grained attributes of each semantic label into our framework. Finally, we will build a website to provide a user interface so that any user can upload his/her own photo, and we output the parsing result within one second. Our framework can also be easily extended to improve the generic image parsing (e.g. scene parsing or human pose estimation) by utilizing the area-specific active templates.

\vspace{-3mm}
\section*{Acknowledgements}

This work is supported by National Natural Science Foundation of China (No. 61328205), the Microsoft Research Asia collaboration projects, Guangdong Natural Science Foundation (no. S2013050014548), Program of Guangzhou Zhujiang Star of Science and Technology (no.2013J2200067), Special Project on Integration of Industry, Educationand Research of Guangdong (no.2012B091000101). 

\vspace{-3mm}
\bibliographystyle{plain}
\scriptsize
\bibliography{clothes}
\vspace{-12mm}

\begin{IEEEbiography}[{\includegraphics[width=1in,height=1.25in,clip,keepaspectratio]{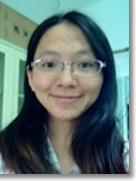}}]{Xiaodan Liang} is a Ph.D. student from School of Information Science and Technology, Sun Yat-sen University, China. She is currently working at National University of Singapore as a Research Intern. Her research interests mainly include semantic segmentation, object/action recognition and medical image analysis.
\end{IEEEbiography}
\vspace{-8ex}
\begin{IEEEbiography}[{\includegraphics[width=1in,height=1.25in,clip,keepaspectratio]{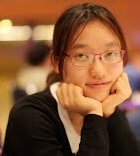}}]{Si Liu} 
is an Associate Professor in Institute of Information Engineering, Chinese Academy of Sciences. She used to be a research fellow at Learning and Vision Group of National University of Singapore. She received her Ph.D. degree from National Laboratory
of Pattern Recognition, Institute of Automation, Chinese Academy of Sciences in 2012. Her research interests includes computer vision and multimedia.
\end{IEEEbiography}
\vspace{-8ex}
\begin{IEEEbiography}[{\includegraphics[width=1in,height=1.25in,clip,keepaspectratio]{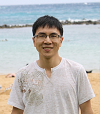}}]{Xiaohui Shen} received his PhD degree from the Department of EECS at Northwestern University in 2013. Before that, he received the MS and BS degrees from the Department of Automation at Tsinghua University, China. He is currently a research scientist at Adobe Research, San Jose, CA. His research
interests include image/video processing and computer vision.
\end{IEEEbiography}
\vspace{-8ex}
\begin{IEEEbiography}[{\includegraphics[width=1in,height=1.25in,clip,keepaspectratio]{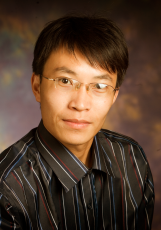}}]{Jianchao Yang}(S’08, M’12) received the M.S. and Ph.D. degrees in electrical and computer engineering from the University of Illinois at Urbana-Champaign, Urbana, in 2011. He is currently a Research Scientist with the Advanced Technology Laboratory, Adobe Systems Inc., San Jose, CA. His research interests include object recognition, deep learning, sparse coding, image/video enhancement, and deblurring.  
\end{IEEEbiography}
\vspace{-8ex}
\begin{IEEEbiography}[{\includegraphics[width=1in,height=1.25in,clip,keepaspectratio]{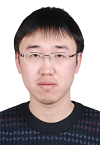}}]{Luoqi Liu} is currently working toward the Ph.D. degree with the Department of Electrical and Computer Engineering, National University of Singapore, Singapore. His research interests include computer vision, multimedia and machine learning. 
\end{IEEEbiography}
\vspace{-8ex}
\begin{IEEEbiography}[{\includegraphics[width=1in,height=1.25in,clip,keepaspectratio]{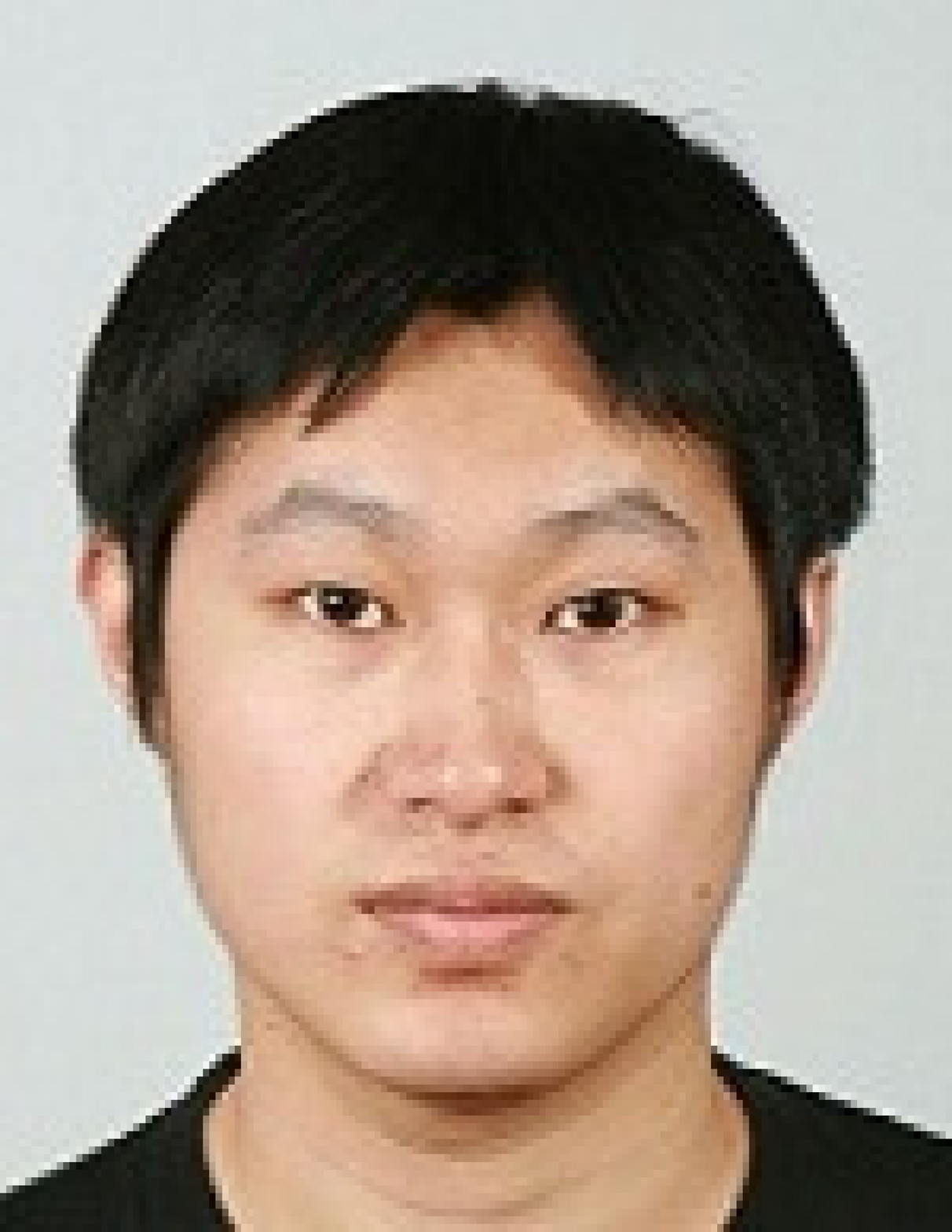}}]{Jian Dong} is now a Research Scientist of Amazon, Seattle. He got his B.Sc degree from University of Science and Technology of China in 2010; and his Ph.D. degree from the National University of Singapore in 2014. His research interests include computer vision and machine learning. He received the winner prizes of classification and segmentation tasks in PASCAL VOC'12, the winner prize of detection task in ImageNet 2014.
\end{IEEEbiography}

\vspace{-12ex}
\begin{IEEEbiography}[{\includegraphics[width=1in,height=1.25in,clip,keepaspectratio]{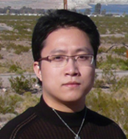}}]{Liang Lin} is a Professor with the School of Advanced Computing, Sun Yat-sen University (SYSU), China. He received the Ph.D. degrees from the Beijing Institute of Technology,in 2008. His Ph.D. dissertation was nominated by the China National Excellent PhD Thesis Award in 2010. He was a Post-Doctoral Research Fellow with the Center for Vision, Cognition, Learning, and Art of UCLA. His research focuses on new models, algorithms and systems for intelligent processing and understanding of visual data. He has published more than 60 papers in top tier academic journals and conferences, and has served as an associate editor for journal Neurocomputing and The Visual Computer. He was supported by several promotive programs or funds for his works, such as “Program for New Century Excellent Talents” of Ministry of Education (China) in 2012. He received the Best Paper Runners-Up Award in ACM NPAR 2010, Google Faculty Award in 2012, and Best Student Paper Award in IEEE ICME 2014. 
\end{IEEEbiography}
\vspace{-10ex}
\begin{IEEEbiography}[{\includegraphics[width=1in,height=1.25in,clip,keepaspectratio]{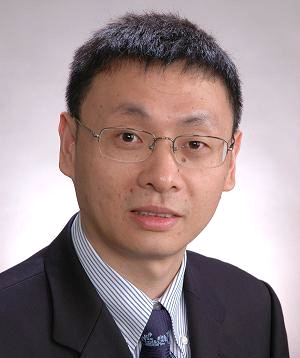}}]{Shuicheng Yan} (M'06-SM'09) is currently an Associate Professor at the Department of Electrical and Computer Engineering at National University of Singapore, and the founding lead of the Learning and Vision Research Group (http://www.lv-nus.org). Dr. Yan's research areas include machine learning, computer vision and multimedia, and he has authored/co-authored nearly 400 technical papers over a wide range of research topics, with Google Scholar citation$>$12,000 times. He is ISI highly-cited researcher 2014, and IAPR Fellow 2014. He has been serving as an associate editor of IEEE TKDE, CVIU and TCSVT. He received the Best Paper Awards from ACM MM'13 (Best Paper and Best Student Paper), ACM MM’12 (Best Demo), PCM'11, ACM MM’10, ICME’10 and ICIMCS'09, the runner-up prize of ILSVRC'13, the winner prizes of the classification task in PASCAL VOC 2010-2012, the winner prize of the segmentation task in PASCAL VOC 2012, the honorable mention prize of the detection task in PASCAL VOC'10, 2010 TCSVT Best Associate Editor (BAE) Award, 2010 Young Faculty Research Award, 2011 Singapore Young Scientist Award, and 2012 NUS Young Researcher Award.
\end{IEEEbiography}

\end{document}